# Neural Network Approach to Stochastic Dynamics for Smooth Multimodal Density Estimation

Z. Zarezadeh
N. Zarezadeh
zakarya.zarezadeh@gmail.com
naeim.zarezade@gmail.com

*Abstract—* In this paper we consider a new probability sampling methods based on Langevin diffusion dynamics to resolve the problem of existing Monte Carlo algorithms when draw samples from high dimensional target densities. We extent Metropolis-Adjusted Langevin Diffusion algorithm by modelling the stochasticity of precondition matrix as a random matrix. An advantage compared to other proposal method is that it only requires the gradient of log-posterior. The proposed method provides fully adaptation mechanisms to tune proposal densities to exploits and adapts the geometry of local structures of statistical models. We clarify the benefits of the new proposal by modelling a Quantum Probability Density Functions of a free particle in a plane (energy Eigen-functions). The proposed model represents a remarkable improvement in terms of performance accuracy and computational time over standard MCMC method.

*Keywords: Neural Network, Stochastic Dynamics, Nonlinear Differential Equations*

## I. INTRODUCTION

Probability distributions over many variables occur frequently in Bayesian inference, statistical physics and simulation studies [1,10-13]. Computational methods for dealing with these large and complex distributions remain an active area of research. As a canonical example, we consider a statistical model $\theta \to x$ for data $x$ generated using parameters $\theta$. The predictive distribution over new data $x^{N+1}$ given observations of N previous settings $\{x^{(n)}\}_{n=1}^{N}$ is an average under the posterior distribution $p\left(\theta \big| \{x^{(n)}\}_{n=1}^{N}\right)$.

$$p(x^{N+1})|\{x^{(n)}\}_{n=1}^{N} = \int p(x^{N+1}|\theta)\, p\left(\theta \big| \{x^{(n)}\}_{n=1}^{N}\right) d\theta$$
$$= \mathbb{E}_{p\left(\theta \big| \{x^{(n)}\}_{n=1}^{N}\right)}[p(x^{N+1})|\theta]$$

where the posterior distribution is given by Bayes' rule

$$p\left(\theta \big| \{x^{(n)}\}_{n=1}^{N}\right) = \frac{p\left(\theta \big| \{x^{(n)}\}_{n=1}^{N} | \theta\right) p(\theta)}{\int p\left(\theta \big| \{x^{(n)}\}_{n=1}^{N} | \theta'\right) p(\theta') d\theta'}$$

Given a parameter space $Q$, we can specify the target distribution as a smooth probability density function, $p(\theta)$, while expectations reduce to integrals over parameter space $\mathbb{E}_p[f] = \int d\theta\, p(\theta) f(\theta)$. Unfortunately, we will not be able to evaluate these integrals analytically for any nontrivial target distribution. The predominant methodology for sampling from such a probability density is Markov Chain Monte Carlo (MCMC) [2,12]. Despite the potential efficiency gains to be obtained in MCMC sampling, the tuning of the MCMC methods remains a major issue especially for challenging inference problems. This paper seeks to address these issues by proposing an adopting method based on Langevin diffusion algorithm for the overall development of MCMC methods. Major steps forward in this regard were made in a proposal process based on the gradient information of the target density. A brief review of proposed method is provided in the following section, and general concepts employed in the study that takes into account the gradient flows of target density are considered. Finally, the new methodology is demonstrated and assessed on a number of interesting statistical problems.

## II. ADAPTIVE PROPOSAL DISTRIBUTION

For an un-normalised probability density function $\tilde{p}(\theta)$, where $\theta \in \mathbb{R}^d$, the normalised density follows as $p(\theta) = \tilde{p}(\theta) / \int \tilde{p}(\theta)\, d\theta$, which is analytically intractable for many statistical models. Monte Carlo estimation of integrals with respect to $p(\theta)$ are therefore required. The predominant methodology for sampling from such a probability density is Markov chain Monte Carlo (MCMC). Consider that $p(\theta)$ denote a probability density function on $\mathbb{R}^d$, from which we desire to draw an ensemble of independent and identically distributed samples. We consider the Langevin diffusion $\dot{\theta} = \nabla_\theta \log(p(\theta)) + \sqrt{2}W$ driven by the time derivative of a standard Brownian motion $W$. In the limit as $t \to \infty$, the probability distribution $p(\theta)$ approaches a stationary distribution, which is also invariant under the diffusion: in fact, it turns out that $p_\infty = p(\theta)$. If we consider the random vector $\theta \in \mathbb{R}^d$ with density $p(\theta)$ and the log density denoted as $L(\theta) \equiv \log(p(\theta))$, then the Metropolis Adjusted

Langevin Algorithm (MALA) [3] is based on a Langevin diffusion, with stationary distribution $p(\theta)$, defined by the stochastic differential equation (SDE)

$$p(\theta(t)) = \frac{1}{2}\nabla_\theta L(\theta(t))dt + db(t)$$

where $b$ denotes a $n$-dimensional Brownian motion [4]. A first-order Euler discretization of the SDE gives the following proposal mechanism

$$\theta^{n+1} = \theta^n + \frac{\varepsilon^2}{2}L(\theta^n) + \varepsilon z^n$$

where $z$ comes from a normal distribution $\mathcal{N}(z|0,I)$ and $\varepsilon$ is the integration step size. An apparent problem with this method is that the resulting proposal is no longer a Markov process (resulting in a non-standard MCMC) and the convergence to the invariant distribution, $p(\theta)$, is no longer guaranteed for finite step size $\varepsilon$, due to the first-order integration error. Instead, it is still possible to obtain a valid algorithm by viewing the chain as a d-dimensional Markov process. This discrepancy can be corrected by employing a Metropolis acceptance probability after each integration step, thus ensuring convergence to the invariant measure. We would like to mention that the isotropic diffusion will be inefficient to draw an independent and identically distributed samples from target density without correlation. MALA incorporates an independent draw from a isotropic multivariate normal distribution on $\mathbb{R}^d$ with mean denoting $\mu(\theta^n, \varepsilon) = \theta^n + \frac{\varepsilon^2}{2}L(\theta^n)$, then the discrete form of the SDE defines a proposal density $q(\theta^*|\theta^n) = \mathcal{N}(\theta^*|\mu(\theta^n,\varepsilon),\varepsilon^2 I)$ with acceptance probability of $\min[1,p(\theta^*)q(\theta^n|\theta^*)/p(\theta^n)q(\theta^*|\theta^n)]$ and covariance matrix equal to $d \times d$ identity matrix scaled by step-size $\varepsilon$ [3]. In contrast to standard MALA algorithm, we would like to have an adaptive proposal mechanism which retain the Markov property. To complete our algorithm, we need to specify a proposal $q$ from which we sample $\theta^*$, and to this aim we make use of the gradient information and the last accepted $\theta$. The way to alleviate this problem, by generating proposals based on drift term (transition kernel), is the following:

$$p(\theta(t)) = \frac{1}{2}\nabla_\theta L(\theta(t))dt + dw(t)$$

$$w(t) = \mathcal{N}(\mu(\theta^n,\varepsilon),\Sigma(\theta^n,\varepsilon))$$

$$\mu(\theta^{n+1},\varepsilon|\theta^n) = \theta^n + \frac{\varepsilon^2}{2}L(\theta^n)$$

$$\Sigma(\theta^n,\varepsilon|\theta^n,\Sigma^n,\nabla_\theta L(\theta^{n-1}))$$
$$= \varepsilon * \frac{\left(\beta + \psi(\Sigma,n) * \left(\left\|\frac{\theta^n}{\theta^{n-1}} - \left(\left\|\frac{\nabla_\theta L(\theta^n)}{\nabla_\theta L(\theta^{n-1})}\right\|^2\right)\right\|\right)^\xi\right)}{1 + \exp\left(-\left\|\frac{\nabla_\theta L(\theta^n)}{\nabla_\theta L(\theta^{n-1})}\right\|^2\right)}$$

$$\psi(\Sigma,n) = \sqrt{2\pi} + \Sigma^n$$

where $\|.\|$ denote the Euclidian norm, $\beta$ is a constant scalar value, $\psi(.)$ is a uniform random variable in the interval of $[0,\sqrt{2\pi}+\Sigma^n]$, and $0 \ll \xi < 1$.

In this manner, we build up the complete path of diffusion, but, as a step size $\varepsilon \to 0$, we recover the solution path of continuous SDE and the target distribution would be the solution of the equation. As mentioned before, $\varepsilon$ is a finite value, then the distribution of the path are biased and Markov properties are modified. We can correct the biases by use of $\theta(\tau+\varepsilon)$ as a proposal mechanism with a mean of $\theta(\tau)$ and the non isotropic covariance, scaled by $\varepsilon$. Therefore, the covariance matrix is encountered by the gradient information of posterior distribution. Up to this stage, we use the directional gradient of posterior distribution to accommodate the small variance. Then, to correct the bias we have to enforce the acceptance probability to ensure detailed balances as following:

$$p(\theta^*|\theta) = \min\left[1,\frac{p(\theta^*)p(\theta|\theta^*)}{p(\theta)p(\theta^*|\theta)}\right]$$

An interesting question is: can geometric structures be employed in proposed methodology? The concept we proposed is not far from the notion of metric tensor and the curvature in Riemannian manifold. Here we have a probability distribution which is completely function of directional derivative of target density. Our proposed method captures the notion of similarities in tangent space defined by metric tensor which is associated with fisher information matrix. So immediately we can see the proposed methodology shared relevant structure with fisher information matrix defined as:

$$X^2(\delta\theta) = \int \frac{|p(y;\theta+\delta\theta) - p(y;\theta)|^2}{p(y;\theta)}dy \approx \delta\theta^T G(\theta)\delta\theta$$

The Fisher information may also be written as:

$$\mathcal{I}(\theta) = \mathbb{E}\left[\left(\frac{\partial}{\partial\theta}\log f(y;\theta)\right)^2 |\theta\right]$$
$$= \int \left(\frac{\partial}{\partial\theta}\log f(y;\theta)\right)^2 f(y;\theta)dy$$

The fisher information has a lot of impact on a target density $f(.)$ and define the overall geometry of space.

*Numerical Simulations:* In this section, we evaluate the performance of our method, which involves sampling from probability density function of a free particle in a plane of dimension $L_x$ and $L_y$. The time-independent Schrödinger equation for this system, as a wave function $|\psi\rangle$, can be written as [5]:

$$-\frac{\hbar^2}{2m}\left(\frac{\partial^2 \psi}{\partial x^2} + \frac{\partial^2 \psi}{\partial y^2}\right) = E\psi$$

The permitted energy values are

$$E_{n_x,n_y} = \frac{\rho^2 \hbar^2}{2m}\left(\frac{n_x^2}{L_x^2} + \frac{n_y^2}{L_y^2}\right)$$

Where $\rho$ is a momentum and $\hbar$ is reduced Planck constant. while the normalised wave function is defined as:

$$\psi_{n_x,n_y}(x,y) = \frac{2}{\sqrt{L_x L_y}} \sin\left(\frac{n_x \pi x}{L_x}\right) \sin\left(\frac{n_y \pi y}{L_y}\right)$$

where $n_x, n_y = 1,2,3,\ldots$. We evaluate the performances of proposed method by comparing it to the pure Hamiltonian Monte Carlo (HMC) [6,7,8] and classic Metropolis-Adjusted Langevin Algorithm (MALA) [3]. In order to evaluate the performance of proposed model, we obtained N=50000 samples from our proposed method, Hamiltonian Monte Carlo and MALA. In Fig. 1, we present the posterior estimates from our method.

around the modes and results in a more efficient exploration of the posterior probability density (Fig. 2). Table 1 demonstrates the overall performance of proposed method in comparison with other two algorithms. We also would like to mention that for numerical simulations we use pseudo-random number generator armed with special features and carefully designed for use in high performance computation based on cellular automata [9,13].

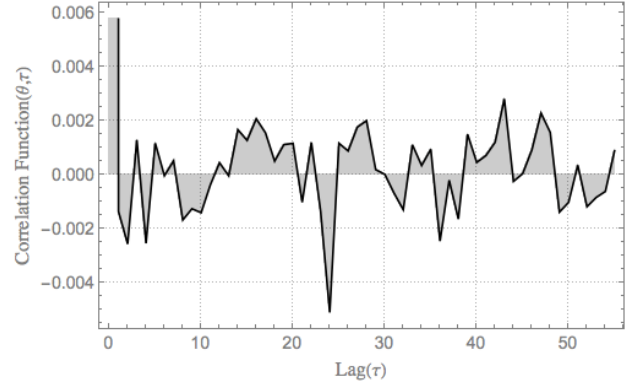

**Fig. 2** The Correlation function for our benchmark test which provides a useful way of characterizing the fast convergence to highly dense local structures.

**Table 1:** Comparison of sampling methods with N=50000 samples.

| Method | Mean Time (s) | Speed |
|---|---|---|
| HMC | No Convergence | - |
| MALA | No Convergence | - |
| Proposed Model | 115 | ×10 |

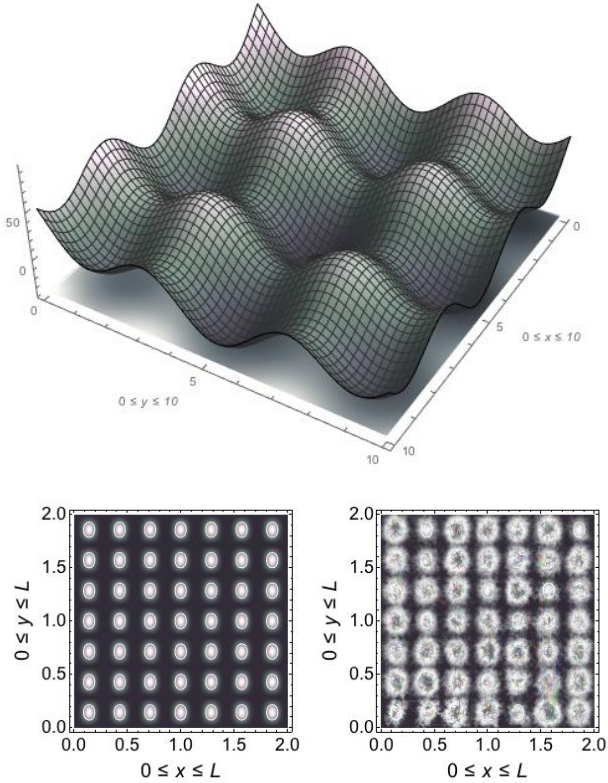

**Fig. 1.** Sampling trajectory of proposed method, indicating a robust mixing property and fast convergence to the theoretical reference values.
(Top)    Target Probability distribution Wave Function
(Left): Target density Contour Plot
(Right): Contour Plot of Estimated Posterior Probability Distribution

It is interesting to conclude that HMC and MALA have failed in the benchmark test because of the complexity of spaces where modes are isolated, small and hard to hit. Instead, the proposed method defines precisely high-density regions

### III. DISCUSSION AND CONCLUSION

In this paper we proposed a new sampling method based on Langevin diffusion dynamics with a random metric, to resolve the problem of existing Monte Carlo algorithms. Therefore, we evaluated it to improve upon existing MCMC methodology in sampling from high dimensional target densities. We provide a probabilistic approach to model stochastic diffusion processes. Through experiments, we visualize the estimated densities between most probable regions, which leads to smoother and more coherent results in contrast to other Monte Carlo sampling methods.